\documentclass[runningheads]{llncs}
\usepackage{graphicx}
\usepackage{hyperref}

\DeclareUrlCommand\ULurl{%
  }

\usepackage{subcaption}
\usepackage{pifont}
\usepackage{booktabs}
\usepackage{multirow}
\usepackage{comment}
\usepackage{pifont}

\newcommand{\xmark}{\ding{55}}

\usepackage[symbol]{footmisc}

\begin{document}
\title{Combining Transformer Generators with Convolutional Discriminators}

%
%

\author{Ricard Durall\thanks{equal contribution}\inst{1,2,3}\ \and
Stanislav Frolov$^*$\inst{4,5} \and
J\"orn Hees\inst{5} \and
Federico Raue\inst{5} \and \\
Franz-Josef Pfreundt \inst{1} \and
Andreas Dengel \inst{4,5}  \and
Janis Keuper\inst{1,6}}
\authorrunning{R. Durall and S. Frolov et al.}
%
\institute{Fraunhofer ITWM, Germany \and
IWR, University of Heidelberg, Germany \and
Fraunhofer Center Machine Learning, Germany \and
Technical University of Kaiserslautern, Germany  \and
German Research Center for Artificial Intelligence (DFKI), Germany \and 
Institute for Machine Learning and Analytics, Offenburg University, Germany
}
\maketitle              
\begin{abstract}
Transformer models have recently attracted much interest from computer vision researchers and have since been successfully employed for several problems traditionally addressed with convolutional neural networks.
At the same time, image synthesis using generative adversarial networks (GANs) has drastically improved over the last few years.
The recently proposed TransGAN is the first GAN using only transformer-based architectures and achieves competitive results when compared to convolutional GANs.
However, since transformers are data-hungry architectures, TransGAN requires data augmentation, an auxiliary super-resolution task during training, and a masking prior to guide the self-attention mechanism.
In this paper, we study the combination of a transformer-based generator and convolutional discriminator and successfully remove the need of the aforementioned required design choices.
We evaluate our approach by conducting a benchmark of well-known CNN discriminators, ablate the size of the transformer-based generator, and show that combining both architectural elements into a hybrid model leads to better results.
Furthermore, we investigate the frequency spectrum properties of generated images and observe that our model retains the benefits of an attention based generator.

\end{abstract}
\section{Introduction}
Generative adversarial networks (GANs) \cite{goodfellow2014generative}, a framework consisting of two neural networks that play a minimax game, made it possible to train generative models for image synthesis in an unsupervised manner.
It consists of a generator network that learns to produce realistic images, and a discriminator network that seeks to discern between real and generated images.
This framework has since successfully been applied to numerous applications such as (unconditional and conditional) image synthesis \cite{karras2019style,brock2018large,frolov2021attrlostgan}, image editing \cite{lee2020maskgan}, text-to-image translation \cite{reed2016generative,frolov2021adversarial}, image-to-image translation \cite{isola2017image}, image super-resolution \cite{ledig2017photo}, and  representation learning \cite{radford2015unsupervised}.
Given the breakthroughs of deep learning enabled by convolutional neural networks (CNNs), GANs typically consist of CNN layers.

While CNNs have been the gold standard in the computer vision community, natural language processing (NLP) problems have recently been dominated by transformer-based architectures \cite{vaswani2017attention}.
On a high level, transformer models consist of an encoder and decoder built from multiple self-attention heads and processes sequences of embedded (word) tokens.
Due to their simple and general architecture, transformers are less restricted by inductive biases and hence well-suited to become universal models.
Inspired by the developments in the field of NLP, researchers have started to apply transformers on image problems by representing an image as a sequence of image patches or pixels \cite{parmar2018image,chen2020generative}.
Since then, there have been major developments of using vision transformers for computer vision applications \cite{carion2020end,dosovitskiy2020image,jiang2021transgan}.

Given the success of adversarial training frameworks, the authors of \cite{jiang2021transgan} conducted the first pilot study to investigate whether a GAN can be created purely from transformer-based architectures to generate images of similar quality.
Their method, termed TransGAN, consists of a generator that progressively increases the feature resolution while decreasing the embedding dimension, and a patch-level discriminator \cite{dosovitskiy2020image}.
However, the discriminator of TransGAN requires careful design choices to reach competitive performance because it seems to be an inferior counterpart unable to provide useful learning signals to the generator.
Given that transformers typically require large datasets, the authors partially alleviate this issue through data augmentation.
Furthermore, they introduce an auxiliary super-resolution task and construct a gradually increasing mask to limit the receptive field of the self-attention mechanism.

\begin{figure}[htb]
\centering
    \includegraphics[width=\linewidth]{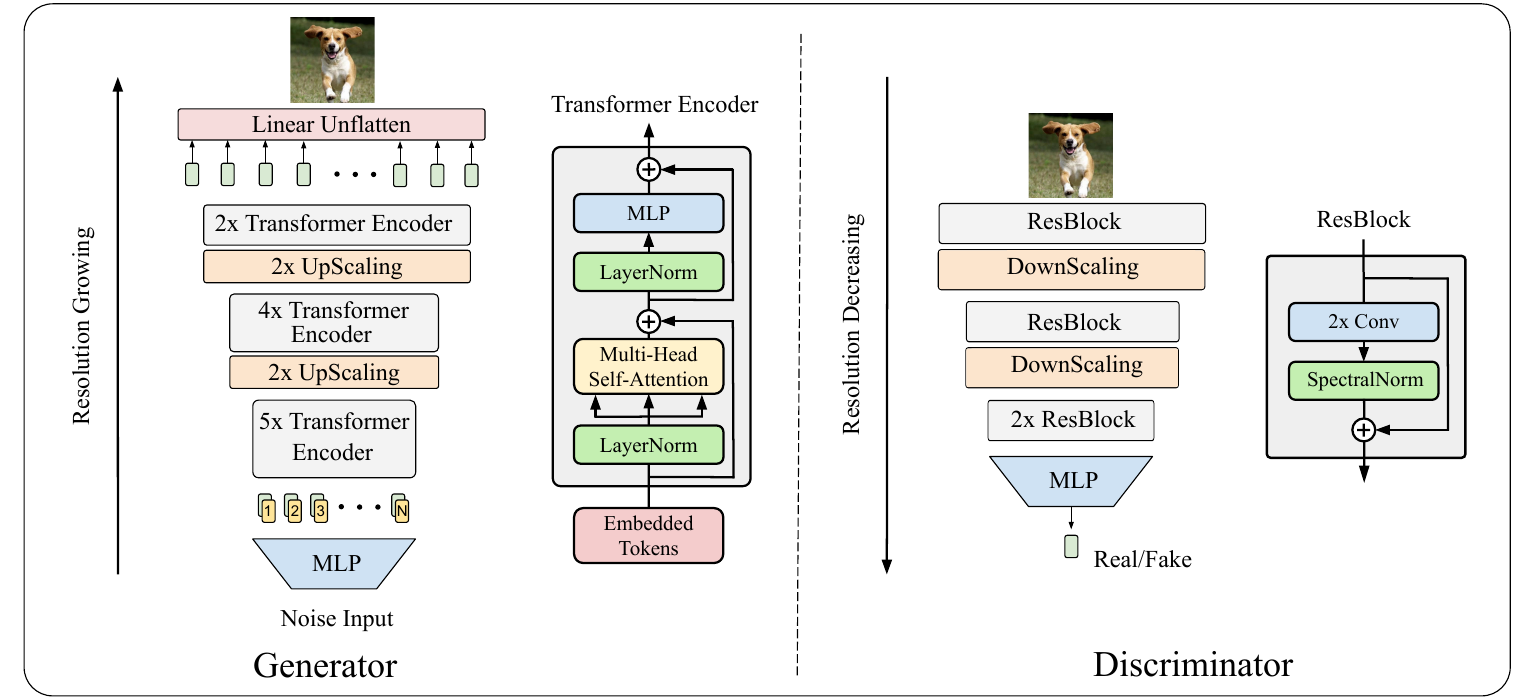}
    \caption{Illustration of our proposed hybrid model consisting of a transformer-based generator and convolutional discriminator.
    While the transformer consists of multiple up-sampling stages combined with transformer encoder blocks, the discriminator consists of down-sampling stages combined with ResBlocks. }
    \label{fig:pipeline}
\end{figure}

Although TransGAN achieves good results and is less restricted by inductive biases, the required design choices are cumbersome.
On the other hand, CNNs have strong biases towards feature locality and spatial invariance induced by the convolutional layers which make them very efficient for problems in the image domain.
Given the success of CNNs for vision problems, in this work we explore the combination of a purely transformer-based generator and CNN discriminator into a hybrid GAN for image synthesis.
In particular, we show that the discriminator of SNGAN \cite{miyato2018spectral} is especially well-suited and leads to improved results on CIFAR-10.
Moreover, our analysis of the frequency spectrum of generated images indicates that our model retains the benefits of a transformer-based generator.
\autoref{fig:pipeline} shows an illustration of our method.
In summary, our contributions are:

\begin{itemize}
    \item we combine a purely transformer-based generator and convolutional discriminator into a hybrid GAN thereby achieving better results on CIFAR-10;
    \item we benchmark several convolutional discriminators, ablate the size of the generator and show that our hybrid method is more robust;
    \item we analyze the frequency spectrum of generated images and show that our hybrid model retains the benefits of the attention-based generator.
\end{itemize}

\section{Related Work}

The goal of generative models is to approximate a real data distribution with a generated data distribution.
To that end, the model is trained to automatically discover and learn the regularities and patterns in the real data assigning a probability to each possible combination of these features.
Eventually, this procedure will encourage the system to mimic the real distribution.
Until recently, most generative models for image synthesis were exclusively built using convolutional layers. 
However, with the uprising of transformers, new topologies started to break the convolution hegemony. 

\subsection{Generative Models Using CNNs}

Generative adversarial network (GAN) \cite{goodfellow2014generative} is one of the most successful generative framework based on convolutional layers. 
They are trained to minimize the distance between the real and generated distributions by optimizing the Jensen-Shannon divergence.
From a game theoretical point of view, this optimization problem can be seen as a minimax game between two players, represented by the discriminator and generator model.
While the generator is trained to generate plausible data, the discriminator's goal is to distinguish between generated and real data, and to penalize the generator for producing unrealistic results.

Variational Autoencoder (VAE) \cite{kingma2013auto} is another very popular framework to train generative models.
Unlike GANs, VAEs explicitly estimate the probability density function of real data by minimizing the Kullback-Leibler (KL) divergence between the two probability distributions.
Similar to an autoencoder, VAEs consist of an encoder and decoder.
The encoder maps the input (e.g., an image) to a latent representation, which usually has a lower dimensionality, to fit a predefined probability distribution.
The decoder is trained to reconstruct the input from the latent representation thereby approximating the original probability distribution.
Once the training has converged to a stable solution, one can sample from the predefined distribution to create new samples.

\subsection{Generative Models Using Attention}
As previously mentioned, there is an increasing tendency pushing attention mechanisms towards visual tasks. 
Image Transformer \cite{parmar2018image} is one of the first approaches to generalize transformers \cite{vaswani2017attention} to a sequence modeling formulation of image generation with a tractable likelihood such as in \cite{van2016pixel,oord2016conditional}.

Another recent approach is ImageGPT \cite{chen2020generative} which was designed to leverage unsupervised representation learning by pre-training a transformer on the image generation task.
As a result, the representations learned by ImageGPT can be used for downstream tasks such as image classification.
The architecture is based on GPT-2 \cite{radford2019language} which trains sequence transformers to auto-regressively predict pixels without incorporating knowledge of the 2D input structure.

Finally, TransGAN \cite{jiang2021transgan} introduced a new GAN paradigm completely free of convolutions and based on pure self-attention blocks.
This transformer-based architecture introduces a novel generator that combines transformer encoders with up-sampling modules consisting of a reshaping and pixelshuffle module \cite{shi2016real} in a multi-level manner.
The discriminator is based on the ViT architecture \cite{dosovitskiy2020image} which was originally developed for image classification without using convolutions.

\subsection{Hybrid Models}
Given the benefits of convolutions and transformers, finding a way to combine both technologies is an interesting and important research question
One successful example of this symbiosis is DALL-E \cite{ramesh2021zero}.
DALL-E is a text-to-image generative model that produces realistic-looking images based on short captions that can specify multiple objects, colors, textures, positions, and other contextual details such as lighting or camera angle.
It has two main blocks: a pre-trained VQ-VAE \cite{oord2017neural} built from convolutional layers which allows the model to generate diverse and high quality images, and a GPT-3 transformer \cite{brown2020language} which learns to map between language and images.
VQGAN \cite{esser2020taming} is a similar architecture but uses images instead of text as input for conditional image synthesis.
GANsformer \cite{hudson2021generative} is a novel approach based on StyleGAN \cite{karras2019style} that employs a bipartite transformer structure coined duplex-attention.
In particular, this model sequentially stacks convolutional layer together with transformer blocks throughout the whole architecture.
This attention mechanism achieves a favorable balance between modeling global phenomena and long-range interaction across the image while maintaining linear efficiency.

\section{Model Architecture}

Our proposed hybrid model is a type of generative model using the GAN framework which involves a generator and discriminator network.
Traditionally, both networks consist of multiple layers of stacked convolutions.
In contrast to previous hybrid models \cite{ramesh2021zero,esser2020taming,hudson2021generative}, in our approach the generator is purely transformer-based and the discriminator only contains convolutions.
See \autoref{fig:pipeline} for an illustration of our model.

\subsection{Transformer Generator}

Originally, transformers were designed for NLP tasks, where they treat individual words as sequential input.
Modelling pixels as individual tokens, even for low-resolution images such as 32$\times$32, is infeasible due to the prohibitive cost (quadratic w.r.t. the sequence length) \cite{jiang2021transgan}.
Inspired by \cite{jiang2021transgan}, we leverage their proposed generator to build our hybrid model.
The generator employs a memory-friendly transformer-based architecture that contains transformer encoders and up-scaling blocks to increase the resolution in a multi-level manner.
The transformer encoder \cite{vaswani2017attention} itself is made of two blocks.
The first is a multi-head self-attention module, while the second is a feed-forward MLP with GELU non-linearity \cite{hendrycks2016gaussian}.
Both blocks use residual connections and layer normalization \cite{ba2016layer}.
Additionally, an up-sampling module after each stage based on pixelshuffle \cite{shi2016real} is inserted.

\subsection{Convolutional Discriminator}
Unlike the generator which has to synthesize each pixel precisely, the discriminator is typically only trained to distinguish between real and fake images.
We benchmark various well-known convolutional discriminators \cite{radford2015unsupervised,karras2020analyzing,zhang2019self,miyato2018spectral} and find that the discriminator of SNGAN \cite{miyato2018spectral} performs particularly well in combination with the transformer-based generator.
It consists of residual blocks (ResBlock) followed by down-sampling layers using average pooling.
The ResBlock itself consists of multiple convolutional layers stacked successively with residual connections, spectral layer normalization \cite{miyato2018spectral} and ReLU non-linearity \cite{nair2010rectified}.

\section{Experiments}

We first provide a detailed description of the experimental setup.
Next, we benchmark various discriminators to investigate their influence on the final performance on the CIFAR-10 dataset.
Then, we ablate the size of the transformer-based generator to assess the effect of the generator's capacity.
Next, we train our proposed method on other datasets and compare with fully convolutional and fully transformer-based architectures.
Finally, we analyze the impact of our proposed method on the frequency spectrum of generated images.

\subsection{Setup}

\noindent \textbf{Models:}
For our empirical investigations, we carry out a benchmark and analyse the impact of different discriminator architectures: DCGAN \cite{radford2015unsupervised}, SNGAN \cite{miyato2018spectral}, SAGAN \cite{zhang2019self}, AutoGAN \cite{gong2019autogan}, StyleGANv2 \cite{karras2020analyzing} and TransGAN.
Furthermore, we  ablate  the  size  of  the transformer-based generator using the scaled-up models of TransGAN \cite{jiang2021transgan} by varying the dimension of the input embedding and/or the number of transformer encoder blocks in each stage.
We built upon the existing codebase\footnote{\href{https://github.com/VITA-Group/TransGAN/tree/7e5fa2d}{https://github.com/VITA-Group/TransGAN/tree/7e5fa2d}} from \cite{jiang2021transgan} with default training procedure and hyperparamters.

\noindent \textbf{Datasets:}
We train our models on four commonly used datasets: CIFAR-10 \cite{krizhevsky2009learning}, CIFAR-100 \cite{krizhevsky2009learning}, STL-10 \cite{coates2011analysis} resized to 48$\times$48, and tiny ImageNet \cite{deng2009imagenet} resized to 32$\times$32.

\noindent\textbf{Metrics:} 
The two most common evaluation metrics are Inception Score (IS) \cite{salimans2016improved} and Fréchet Inception Distance (FID) \cite{heusel2017gans}.
While IS computes the KL divergence between the conditional class distribution and the marginal class distribution over the generated data, FID calculates Fréchet distance between multivariate Gaussian fitted to the intermediate activations of the Inception-v3 network \cite{szegedy2016rethinking} of generated and real images.

\subsection{Results}
We first study the role and influence of the discriminator topology on the final performance using the CIFAR-10 dataset.
To that end, the generator architecture remains fixed, while the discriminator architecture is swapped with various standard CNN discriminators.
\autoref{tab:cifar10_benchmark} contains the scores for combination, where we can see that the discriminator of SNGAN leads to the best IS and FID and, unlike TransGAN, our model does not require data augmentation, auxiliary tasks nor any kind of locality-aware initialization for the attention blocks.
However, we can observe an influence of the number of parameters on the final results.
If the discriminator is strong, such as in StyleGANv2 \cite{karras2020analyzing}, our transformer-based generator is not able to improve as quickly and consequently results in poor performance.
If the discriminator is too small, such as in DCGAN \cite{radford2015unsupervised}, it will not be able to provide good learning signals to the generator.
Finally, we ablate the impact of Spectral Normalization (SN) on the discriminator and observe just slightly worse results which indicates that SN is not the main contributor to the good overall performance of this combination.

\begin{table}[!ht]
\centering
\caption{
Benchmark results on CIFAR-10 using different discriminator architectures.
\xmark{} indicates unavailable scores due to a collapsed model during training.
Using the SNGAN discriminator, we can achieve better scores without the need of data augmentation, auxilifoldary tasks and mask guidance.
}
\label{tab:cifar10_benchmark}
\begin{tabular}{lrrr}
\toprule
Discriminator & Params. (M) & IS $\uparrow$ & FID $\downarrow$  \\
\midrule
DCGAN \cite{radford2015unsupervised} & 0.6 & \xmark & \xmark  \\
StyleGANv2 \cite{karras2020analyzing} & 21.5 & 4.19 & 127.25 \\
SAGAN \cite{zhang2019self} & 1.1 & 7.29 & 26.08 \\
AutoGAN \cite{gong2019autogan} & 9.4 & 8.59 & 13.23 \\
TransGAN \cite{jiang2021transgan} & 12.4 & 8.63 & 11.89\\
SNGAN w/o SN & 9.4 & 8.79 & 9.45 \\
SNGAN \cite{miyato2018spectral} & 9.4 & \textbf{8.81} & \textbf{8.95}\\
\bottomrule
\end{tabular}
\end{table}

\begin{table}[!ht]
\centering
\caption{
Benchmark results on CIFAR-10 using different generator sizes together with the convolutional SNGAN discriminator.
Our hybrid model achieves consistently better scores when compared to the full transformer-based GAN, especially for small transformer-based generators. 
Note that our approach does not employ any additional mechanisms during training.
}
\label{tab:cifar10_benchmark_generator}
\begin{tabular}{lrlrr}
\toprule
Generator  & Params. (M) & Discriminator  & IS $\uparrow$ & FID $\downarrow$ \\
\midrule
\multirow{2}{*}{TransGAN-S} & \multirow{2}{*}{18.6} & TransGAN & 8.22 & 18.58 \\
                            &                        & SNGAN & \textbf{8.79} & \textbf{9.95}\\
\midrule
\multirow{2}{*}{TransGAN-M} & \multirow{2}{*}{33.1} & TransGAN & 8.36 & 16.27 \\
                            &                        & SNGAN & \textbf{8.80} & \textbf{9.53}\\
\midrule
\multirow{2}{*}{TransGAN-L} & \multirow{2}{*}{74.3} & TransGAN & 8.50 & 14.46 \\
                            &                        & SNGAN & \textbf{8.81} & \textbf{8.97}\\
\midrule
\multirow{2}{*}{TransGAN-XL}& \multirow{2}{*}{133.6}& TransGAN & 8.63 & 11.89 \\
                            &                       & SNGAN & \textbf{8.81} & \textbf{8.95}\\
\bottomrule
\end{tabular}
\end{table}

After benchmarking the discriminator topology, we ablate the size of the transformer-based generator using the scaled-up models of \cite{jiang2021transgan}.
\autoref{tab:cifar10_benchmark_generator} contains the scores for each configuration where we can see how our approach systematically outperforms the fully transformer-based version.
Additionally, we observe how a bigger capacity leads to better results, but with marginal gains above TransGAN-L size.
This behaviour indicates a saturation in the generative model and hence adding more capacity might not further improve the results.

\begin{figure}[!t]
\begin{subfigure}{0.24\linewidth}
    \includegraphics[width=\linewidth]{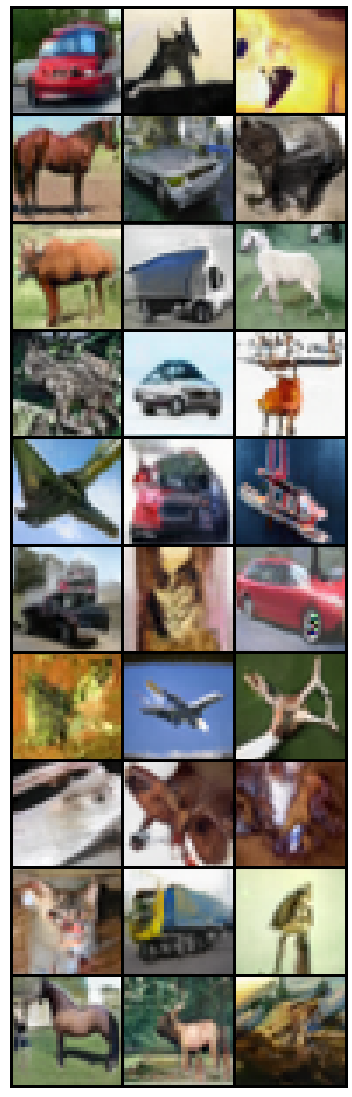}
     \caption{CIFAR-10}
\end{subfigure}
\begin{subfigure}{0.24\linewidth}
    \includegraphics[width=\linewidth]{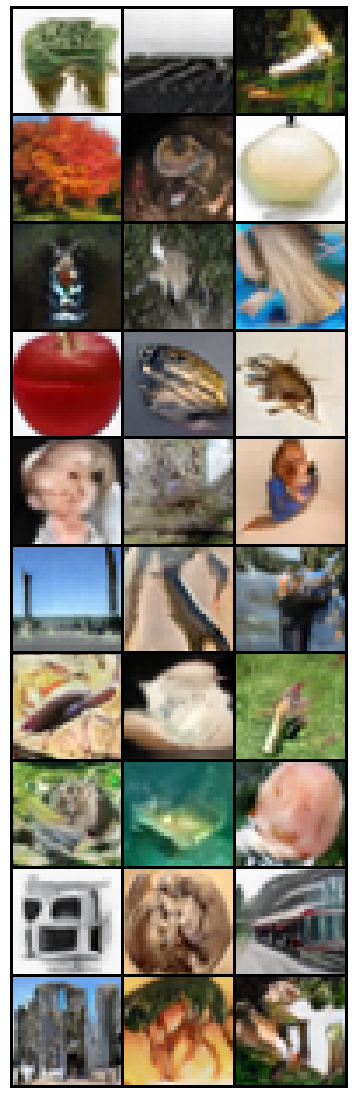}
    \caption{CIFAR-100}
\end{subfigure}
\begin{subfigure}{0.24\linewidth}
    \includegraphics[width=\linewidth]{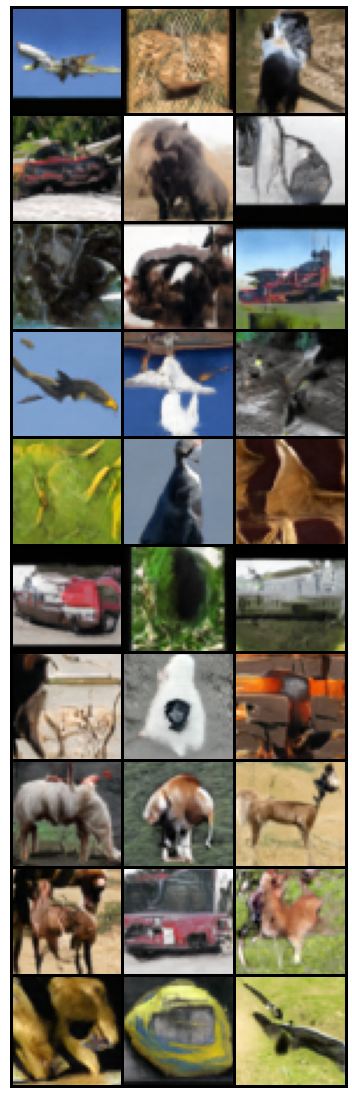}
     \caption{STL-10}
\end{subfigure}
\begin{subfigure}{0.24\linewidth}
    \includegraphics[width=\linewidth]{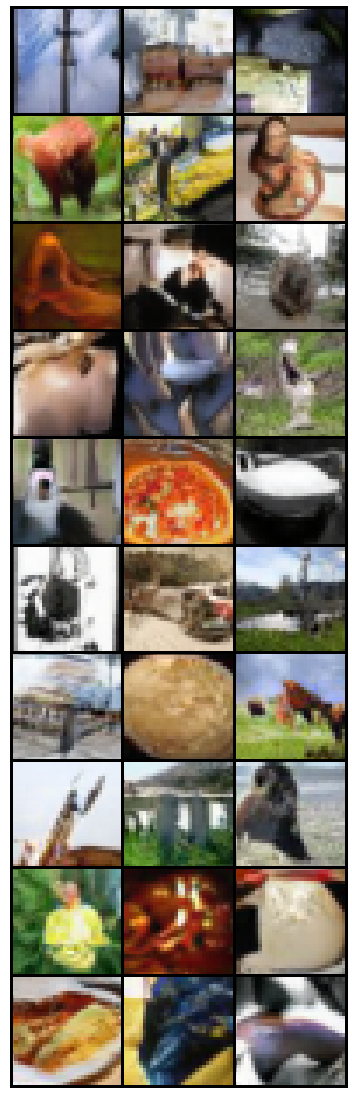}
    \caption{ImageNet}
\end{subfigure}

\caption{
Random generated samples of our method trained on different datasets.
The images are visually pleasing and diverse in shape and in texture.
}
\label{fig:samples}
\end{figure}

\autoref{tab:results_on_other_datasets} contains results on other commonly used datasets.
Our method, consisting of a transformer-based generator and SNGAN discriminator achieved similar or better results without requiring data augmentation, auxiliary tasks or mask guidance.
\autoref{fig:samples} shows random generated samples which appear to be natural, visually pleasing and diverse in shape and in texture.

\begin{table}[t]
\centering
\caption{
FID results on various datasets.
Our hybrid model achieves similar or better scores when compared to either a full convolutional or full transformer-based GAN.
}
\label{tab:results_on_other_datasets}
\begin{tabular}{lrrrr}
\toprule
FID $\downarrow$ & CIFAR-10 & CIFAR-100 & STL-10 & ImageNet \\
\midrule
SNGAN \cite{miyato2018spectral} & 22.16 & 27.13 & 43.75 & 29.30 \\
TransGAN \cite{jiang2021transgan} & 11.89 & - & \textbf{25.32} & - \\
Ours & \textbf{8.95} & \textbf{14.29} & 31.30 & \textbf{14.53} \\
\bottomrule
\end{tabular}
\end{table}

\subsection{Frequency Analysis}
While a good score on the chosen image metric is one way to assess the performance of a given model, there are other, equally important properties, that need to be evaluated.
Recently, \cite{durall2020watch} observed that commonly used convolutional up-sampling operations might lead to the inability to learn the spectral distribution of real images, especially their high-frequency components.
Furthermore, these  artifacts seem to be present in all kinds of convolutional based models, independently of their topology.
Following prior works \cite{durall2019unmasking,durall2020watch}, we also employ the azimuthal integration over the Fourier power spectrum to analyze the spectral properties of generated images, and we extend the evaluation to non-convolutional based systems.
In particular, we conduct experiments on  pure attention, pure convolutional and hybrid architectures trained on CIFAR-10.

\begin{figure}[htb]
\centering
    \includegraphics[width=\linewidth]{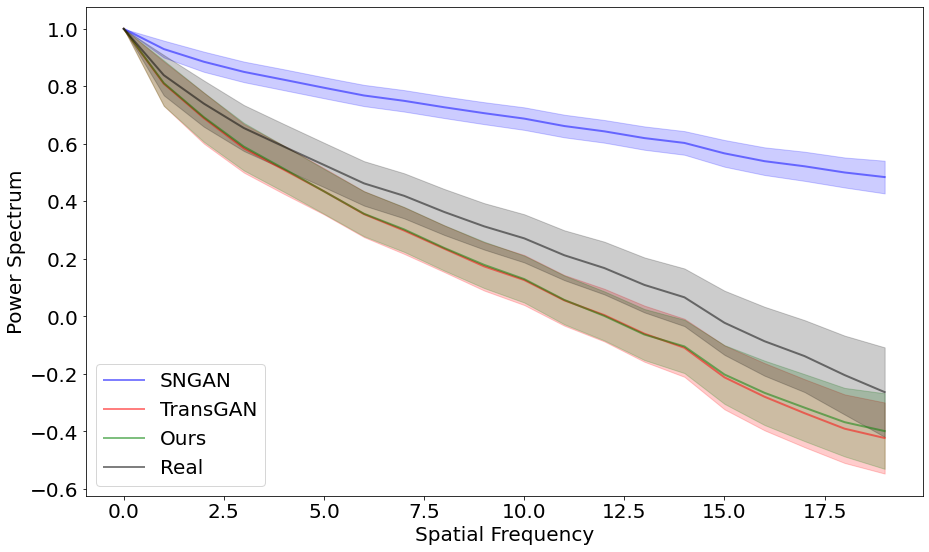}
\caption{
Power spectrum results of real and generated images on CIFAR-10.
Statistics (mean and variance) after azimuthal integration over the power spectrum of real and generated images.
Our hybrid model displays good spectral properties indicated by a response similar to the real data.
}
\label{fig:freq}
\end{figure}

\autoref{fig:freq} displays the power spectrum of real data and images generated by three different models.
Among them, one is based on convolutions (SNGAN), one is based on transformers (TransGAN), and one is a hybrid model (ours).
Notice how the pure CNN approach has a significantly worse power spectrum indicated by an unmatched frequency response when compared to real data.
TransGAN and our method are much more aligned with the real spectrum, but still there is a substantial gap.
By using the transformer-based generator and a strong CNN discriminator, our model achieves better results in terms of IS and FID without the additional data augmentation, auxiliary loss and masking prior while retaining a good frequency spectrum.

\section{Discussion}
Our method successfully combines transformers with convolutions into a hybrid GAN model and achieves similar or better results compared to its fully convolutional and fully attentional baselines.
Furthermore, our method removes the need of data augmentation, auxiliary learning tasks, and masking priors to guide the self-attention mechanism.
Additionally, images generated by our hybrid approach retain the benefits of the attention-based TransGAN in terms of frequency spectrum.

By benchmarking several discriminator topologies and differently-sized generators, we found that the capacity of the convolutional discriminator must be aligned to the capacity of the transformer-based generator and cannot be too big or too small to achieve good performance.
Moreover, our method performs much more reliable across generators of different sizes and consistently achieves better scores.

Even though our method leads to promising results, more work is required to investigate optimal ways to combine transformers and convolutions into strong GANs.
To the best of our knowledge, there are currently only two other GAN approaches that use both transformers and convolutions, GANsformer \cite{hudson2021generative} and VQGAN \cite{esser2020taming}.
However, they have completely different setups. 
While GANsformer and VQGAN integrate self-attention layers in-between the architecture in a sandwich like way, we keep them separated.
In particular, our approach consists of a purely transformer-based generator, and a fully CNN-based discriminator, thereby constraining the interaction between attention and convolutions. 
Hence, our approach maintains relaxed inductive biases that characterize transformers in the generator, while leveraging the useful ones in the discriminator.
Last but not least, our frequency spectrum analysis has brought new insights regarding the impact of transformers on the generated images.
It shows, how a pure transformer based GAN framework, such as in TransGAN \cite{jiang2021transgan}, seems to learn the frequency components in a more accurate manner.
Our hybrid model is able to maintain the well-matched spectrum, while achieving better or similar scores without requiring additional training constraints.
We think that these findings can lead to a new paradigm, where both transformers and convolutions are used to generate images.

\section{Conclusion}

Motivated by the desire to obtain the best from transformers and convolutions, in this work we proposed a hybrid model using a pure transformer-based generator, and a standard convolutional discriminator.
Typically, transformers rely on relaxed inductive biases thereby making them universal models.
As a consequence, they require vast amounts of training data.
However, our method leverages the benefits of convolutions through the discriminator feedback, while retaining the advantages of transformers in the generator.
Our hybrid approach achieves competitive scores across multiple common datasets, and does not need data augmentation, auxiliary learning tasks, and masking priors for the attention layers to successfully train the model.
Additionally, it inherits the well-matched spectral properties from its transformer-based generator baseline.
We hope this approach can pave the way towards new architectural designs, where the benefits of different architectural designs can successfully be combined into one.
Possible future research directions could be investigating the importance of including inductive biases into the architectures of the generator and discriminator, respectively, as well as scaling our hybrid approach to higher resolutions.

\bibliographystyle{splncs04}
\bibliography{samplepaper}

\begin{thebibliography}{10}
\providecommand{\url}[1]{\texttt{#1}}
\providecommand{\urlprefix}{URL }
\providecommand{\doi}[1]{https://doi.org/#1}

\bibitem{ba2016layer}
Ba, J.L., Kiros, J.R., Hinton, G.E.: Layer normalization. arXiv:1607.06450
  (2016)

\bibitem{brock2018large}
Brock, A., Donahue, J., Simonyan, K.: Large scale gan training for high
  fidelity natural image synthesis. In: International Conference on Learning
  Representations (2018)

\bibitem{brown2020language}
Brown, T.B., Mann, B., Ryder, N., Subbiah, M., Kaplan, J., Dhariwal, P.,
  Neelakantan, A., Shyam, P., Sastry, G., Askell, A., et~al.: Language models
  are few-shot learners. In: Advances in Neural Information Processing Systems
  (2020)

\bibitem{carion2020end}
Carion, N., Massa, F., Synnaeve, G., Usunier, N., Kirillov, A., Zagoruyko, S.:
  End-to-end object detection with transformers. In: European Conference on
  Computer Vision. pp. 213--229 (2020)

\bibitem{chen2020generative}
Chen, M., Radford, A., Child, R., Wu, J., Jun, H., Luan, D., Sutskever, I.:
  Generative pretraining from pixels. In: International Conference on Machine
  Learning. pp. 1691--1703 (2020)

\bibitem{coates2011analysis}
Coates, A., Ng, A., Lee, H.: An analysis of single-layer networks in
  unsupervised feature learning. In: Proceedings of the International
  Conference on Artificial Intelligence and Statistics. pp. 215--223 (2011)

\bibitem{deng2009imagenet}
Deng, J., Dong, W., Socher, R., Li, L.J., Li, K., Fei-Fei, L.: Imagenet: A
  large-scale hierarchical image database. In: Proceedings of the IEEE Computer
  Vision and Pattern Recognition. pp. 248--255 (2009)

\bibitem{dosovitskiy2020image}
Dosovitskiy, A., Beyer, L., Kolesnikov, A., Weissenborn, D., Zhai, X.,
  Unterthiner, T., Dehghani, M., Minderer, M., Heigold, G., Gelly, S., et~al.:
  An image is worth 16x16 words: Transformers for image recognition at scale.
  In: International Conference on Learning Representations (2021)

\bibitem{durall2020watch}
Durall, R., Keuper, M., Keuper, J.: Watch your up-convolution: Cnn based
  generative deep neural networks are failing to reproduce spectral
  distributions. In: Proceedings of the IEEE Computer Vision and Pattern
  Recognition. pp. 7890--7899 (2020)

\bibitem{durall2019unmasking}
Durall, R., Keuper, M., Pfreundt, F.J., Keuper, J.: Unmasking deepfakes with
  simple features. arXiv:1911.00686  (2019)

\bibitem{esser2020taming}
Esser, P., Rombach, R., Ommer, B.: Taming transformers for high-resolution
  image synthesis. arXiv:2012.09841  (2020)

\bibitem{frolov2021adversarial}
Frolov, S., Hinz, T., Raue, F., Hees, J., Dengel, A.: Adversarial text-to-image
  synthesis: A review. arXiv:2101.09983  (2021)

\bibitem{frolov2021attrlostgan}
Frolov, S., Sharma, A., Hees, J., Karayil, T., Raue, F., Dengel, A.:
  Attrlostgan: Attribute controlled image synthesis from reconfigurable layout
  and style. arXiv:2103.13722  (2021)

\bibitem{gong2019autogan}
Gong, X., Chang, S., Jiang, Y., Wang, Z.: Autogan: Neural architecture search
  for generative adversarial networks. In: Proceedings of the IEEE
  International Conference on Computer Vision. pp. 3224--3234 (2019)

\bibitem{goodfellow2014generative}
Goodfellow, I.J., Pouget-Abadie, J., Mirza, M., Xu, B., Warde-Farley, D.,
  Ozair, S., Courville, A., Bengio, Y.: Generative adversarial networks. In:
  Advances in Neural Information Processing Systems (2014)

\bibitem{hendrycks2016gaussian}
Hendrycks, D., Gimpel, K.: Gaussian error linear units (gelus).
  arXiv:1606.08415  (2016)

\bibitem{heusel2017gans}
Heusel, M., Ramsauer, H., Unterthiner, T., Nessler, B., Hochreiter, S.: Gans
  trained by a two time-scale update rule converge to a local nash equilibrium.
  In: Advances in Neural Information Processing Systems (2017)

\bibitem{hudson2021generative}
Hudson, D.A., Zitnick, C.L.: Generative adversarial transformers.
  arXiv:2103.01209  (2021)

\bibitem{isola2017image}
Isola, P., Zhu, J.Y., Zhou, T., Efros, A.A.: Image-to-image translation with
  conditional adversarial networks. In: Proceedings of the IEEE Computer Vision
  and Pattern Recognition. pp. 1125--1134 (2017)

\bibitem{jiang2021transgan}
Jiang, Y., Chang, S., Wang, Z.: Transgan: Two transformers can make one strong
  gan. arXiv:2102.07074v2  (2021), \url{"https://arxiv.org/abs/2102.07074v2"}

\bibitem{karras2019style}
Karras, T., Laine, S., Aila, T.: A style-based generator architecture for
  generative adversarial networks. In: Proceedings of the IEEE Computer Vision
  and Pattern Recognition. pp. 4401--4410 (2019)

\bibitem{karras2020analyzing}
Karras, T., Laine, S., Aittala, M., Hellsten, J., Lehtinen, J., Aila, T.:
  Analyzing and improving the image quality of stylegan. In: Proceedings of the
  IEEE Computer Vision and Pattern Recognition. pp. 8110--8119 (2020)

\bibitem{kingma2013auto}
Kingma, D.P., Welling, M.: Auto-encoding variational bayes. In: International
  Conference on Learning Representations (2013)

\bibitem{krizhevsky2009learning}
Krizhevsky, A., Hinton, G., et~al.: Learning multiple layers of features from
  tiny images. Tech. rep., University of Toronto (2009)

\bibitem{ledig2017photo}
Ledig, C., Theis, L., Husz{\'a}r, F., Caballero, J., Cunningham, A., Acosta,
  A., Aitken, A., Tejani, A., Totz, J., Wang, Z., et~al.: Photo-realistic
  single image super-resolution using a generative adversarial network. In:
  Proceedings of the IEEE Computer Vision and Pattern Recognition. pp.
  4681--4690 (2017)

\bibitem{lee2020maskgan}
Lee, C.H., Liu, Z., Wu, L., Luo, P.: Maskgan: Towards diverse and interactive
  facial image manipulation. In: Proceedings of the IEEE Computer Vision and
  Pattern Recognition. pp. 5549--5558 (2020)

\bibitem{miyato2018spectral}
Miyato, T., Kataoka, T., Koyama, M., Yoshida, Y.: Spectral normalization for
  generative adversarial networks. In: International Conference on Learning
  Representations (2018)

\bibitem{nair2010rectified}
Nair, V., Hinton, G.E.: Rectified linear units improve restricted boltzmann
  machines. In: International Conference on Machine Learning (2010)

\bibitem{oord2016conditional}
Oord, A.v.d., Kalchbrenner, N., Vinyals, O., Espeholt, L., Graves, A.,
  Kavukcuoglu, K.: Conditional image generation with pixelcnn decoders. In:
  Advances in Neural Information Processing Systems (2016)

\bibitem{oord2017neural}
Oord, A.v.d., Vinyals, O., Kavukcuoglu, K.: Neural discrete representation
  learning. In: Advances in Neural Information Processing Systems (2017)

\bibitem{parmar2018image}
Parmar, N., Vaswani, A., Uszkoreit, J., Kaiser, L., Shazeer, N., Ku, A., Tran,
  D.: Image transformer. In: International Conference on Machine Learning. pp.
  4055--4064 (2018)

\bibitem{radford2015unsupervised}
Radford, A., Metz, L., Chintala, S.: Unsupervised representation learning with
  deep convolutional generative adversarial networks. In: International
  Conference on Learning Representations (2015)

\bibitem{radford2019language}
Radford, A., Wu, J., Child, R., Luan, D., Amodei, D., Sutskever, I.: Language
  models are unsupervised multitask learners. OpenAI blog  \textbf{1}(8), ~9
  (2019)

\bibitem{ramesh2021zero}
Ramesh, A., Pavlov, M., Goh, G., Gray, S., Voss, C., Radford, A., Chen, M.,
  Sutskever, I.: Zero-shot text-to-image generation. arXiv:2102.12092  (2021)

\bibitem{reed2016generative}
Reed, S., Akata, Z., Yan, X., Logeswaran, L., Schiele, B., Lee, H.: Generative
  adversarial text to image synthesis. In: International Conference on Machine
  Learning. pp. 1060--1069 (2016)

\bibitem{salimans2016improved}
Salimans, T., Goodfellow, I., Zaremba, W., Cheung, V., Radford, A., Chen, X.:
  Improved techniques for training gans. In: Advances in Neural Information
  Processing Systems (2016)

\bibitem{shi2016real}
Shi, W., Caballero, J., Husz{\'a}r, F., Totz, J., Aitken, A.P., Bishop, R.,
  Rueckert, D., Wang, Z.: Real-time single image and video super-resolution
  using an efficient sub-pixel convolutional neural network. In: Proceedings of
  the IEEE Computer Vision and Pattern Recognition. pp. 1874--1883 (2016)

\bibitem{szegedy2016rethinking}
Szegedy, C., Vanhoucke, V., Ioffe, S., Shlens, J., Wojna, Z.: Rethinking the
  inception architecture for computer vision. In: Proceedings of the IEEE
  Computer Vision and Pattern Recognition. pp. 2818--2826 (2016)

\bibitem{van2016pixel}
Van~Oord, A., Kalchbrenner, N., Kavukcuoglu, K.: Pixel recurrent neural
  networks. In: International Conference on Machine Learning. pp. 1747--1756
  (2016)

\bibitem{vaswani2017attention}
Vaswani, A., Shazeer, N., Parmar, N., Uszkoreit, J., Jones, L., Gomez, A.N.,
  Kaiser, L., Polosukhin, I.: Attention is all you need. In: Advances in Neural
  Information Processing Systems. pp. 5998--6008 (2017)

\bibitem{zhang2019self}
Zhang, H., Goodfellow, I., Metaxas, D., Odena, A.: Self-attention generative
  adversarial networks. In: International Conference on Machine Learning. pp.
  7354--7363 (2019)

\end{thebibliography}
\end{document}